\documentclass{article}
\usepackage[final,nonatbib]{neurips_2018}

\usepackage[utf8]{inputenc} 
\usepackage[T1]{fontenc}    
\usepackage{hyperref}       
\usepackage{url}            
\usepackage{booktabs}       
\usepackage{amsfonts}       
\usepackage{nicefrac}       
\usepackage{microtype}      
\usepackage{amsmath,amssymb}
\usepackage{algorithmic}
\usepackage{subcaption}
\usepackage{graphicx}

\usepackage{multirow}
\usepackage[boxruled,linesnumbered]{algorithm2e}

\DeclareMathOperator*{\argmin}{arg\,min}

\newcommand{\Rnx}[2]{\MBB{R}^{#1\times #2}}

\newcommand{\EQ}[2]
{
	\begin{equation}
		{#1}
		{#2}
	\end{equation}
}

\newcommand{\K}{{\mathcal K}}
\newcommand{\MB}[1]{\mathbf{#1}}
\newcommand{\MBB}[1]{\mathbb{#1}}

\newcommand{\MC}[1]{\mathcal{#1}}

\newcommand{\Tr}[0]{\mathrm{Tr}}

\newcommand{\T}{\top}
\newcommand{\Bsum}{\mathlarger\sum}
\newcommand{\1}{\MB{1}}

\newcommand{\smalleqb}[1]
{
	\begingroup
	\makeatletter
	\def
	\f@size{1}
	{#1}
    \endgroup
}

\newcommand{\nf}{d}
\newcommand{\nd}{k}
\newcommand{\ncl}{c}

\newcommand{\JA}{\MC{J}_{A}}

\newcommand{\ny}{{\vec{x}}}
\newcommand{\nY}{\mathbf{X}}
\newcommand{\nxs}{\gamma}
\newcommand{\nx}{{\vec{\gamma}}}
\newcommand{\nX}{\mathbf{\Gamma}}
\newcommand{\nU}{\mathbf{U}}

\newcommand{\nA}{\mathbf{A}}
\newcommand{\na}{{\vec{a}}}
\newcommand{\nas}{a}
\newcommand{\nh}{{\vec{h}}}
\newcommand{\nhs}{h}
\newcommand{\nH}{\mathbf{H}}
\newcommand{\nALF}{\mathbf{B}}
\newcommand{\nAlf}{\vec{\beta}}
\newcommand{\nalf}{{\beta}}
\newcommand{\phih}{\hat{\Phi}}
\newtheorem{defin}{Definition}

\title{Interpretable Discriminative Dimensionality Reduction and Feature Selection on the Manifold}

\author{%
	Babak Hosseini
	\thanks{
		Preprint of the publication~\cite{hosseini2019Interpretable}, as provided by the authors.
		The final publication is available at \url{https://link.springer.com/conference/ecml}
	} \\
	Center of Cognitive Interactive Technology (CITEC)\\
	Bielefeld University, Germany\\
	\texttt{bhosseini@techfak.uni-bielefeld.de} \\
	\And
	Barbara Hammer\\
	Center of Cognitive Interactive Technology (CITEC)\\
	Bielefeld University, Germany\\
	\texttt{bhammer@techfak.uni-bielefeld.de} \\
}

\begin{document}
\maketitle

\begin{abstract}
Dimensionality reduction (DR) on the manifold includes effective methods which project the data from an implicit relational space onto a vectorial space. 
Regardless of the achievements in this area, these algorithms suffer from the lack of interpretation of the projection dimensions. 
Therefore, it is often difficult to explain the physical meaning behind the embedding dimensions. 
In this research, we propose the interpretable kernel DR algorithm (I-KDR) as a new algorithm which maps the data from the feature space to a lower dimensional space where the classes are more condensed with less overlapping. 
Besides, the algorithm creates the dimensions upon local contributions of the data samples, which makes it easier to interpret them by class labels.
Additionally, we efficiently fuse the DR with feature selection task to select the most relevant features of the original space to the discriminative objective. 
Based on the empirical evidence, I-KDR provides better interpretations for embedding dimensions as well as higher discriminative performance in the embedded space compared to the state-of-the-art and popular DR algorithms.
\end{abstract}

\section{Introduction}
Dimensionality reduction (DR) is an essential preprocessing phase in the application of many algorithms in machine learning and data analytics. 
The general goal in any DR approach is to obtain an embedding to transfer the data from the original high-dimensional (HD) space to a low-dimension (LD) space, such that this projection preserves the vital information about the data distribution\cite{roweis2000nonlinear}.
It is common to split the dimensionality reduction methods into two groups of unsupervised and supervised algorithms. 
The first group includes methods such as Principal Component Analysis (PCA)~\cite{jolliffe2011principal} which 
finds a new embedding space in which the dimensions are sorted based on the maximum data variation they can achieve, 
or locally linear embedding (LLE)~\cite{roweis2000nonlinear} 
that focuses on preserving the relational structure of data points in the local neighborhoods of the space throughout an embedding.

The second group of algorithms, known as supervised (discriminative) DR methods, assume that data classes can obtain the same or even better separations in an intrinsic LD space. 
As a popular supervised algorithm, Linear Discriminant Analysis (LDA)~\cite{mika1999fisher} tries to find a mapping which increases the distance between the class centroids while preserving the intra-class variations.
Its subsequent algorithms such as LLDA~\cite{kim2005locally} and CPM~\cite{ye2006cpm} tried to relax the constraints on within-class variations to project the sub-clusters to the LD space more efficiently.

It is possible to consider an implicit mapping of data to a high-dimensional reproducing kernel Hilbert space (RKHS) 
primarily to obtain a relational representation of the non-vectorial or structured data distributions.
Consequently, a branch of DR algorithms (kernel-DR) is focused on kernel-based data representations to transfer the data from the original RKHS to a vectorial space. 
This projection can become significant especially when it makes the application of many vectorial algorithms possible on LD embedding of such data. 
The most famous kernel-DR algorithms are Kernelized PCA (K-PCA)
and K-FDA~\cite{mika1999fisher} which are the kernelized versions of PCA and LDA algorithms respectively.
In these methods and many other kernel-DR algorithms, it is common to construct the embedding dimensions upon different weighted combinations of data points in the original RKHS.
Other notable examples of kernel-based methods include algorithms such as  KDR~\cite{fukumizu2004kernel}, KEDR~\cite{alvarez2017kernel},
and LDR~\cite{suzuki2013sufficient}.

Additionally, by assuming a set of non-linear mappings to different sub-spaces in the feature space, it is possible to obtain one specific kernel representation for each dimension of the data~\cite{dileep2009representation,gonen2011multiple}. 
Consequently, a specific group of methods tried to apply DR frameworks also to feature selection tasks on manifolds~\cite{lin2011multiple,jiang2014trace}.

\begin{figure}
	\centering		
	\begin{subfigure}{0.23\textwidth}
		\includegraphics[width=1\linewidth]{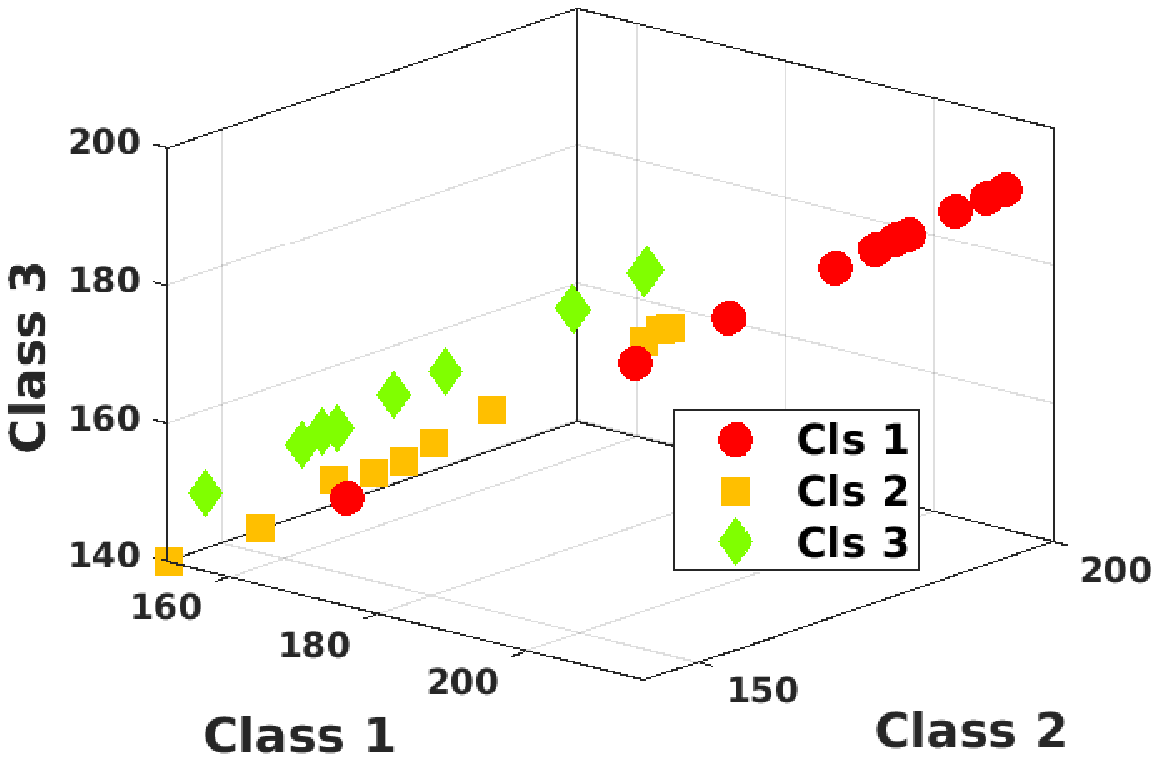}
		\caption{}	
	\end{subfigure}	
	\begin{subfigure}{0.23\textwidth}
		\includegraphics[width=1\linewidth]{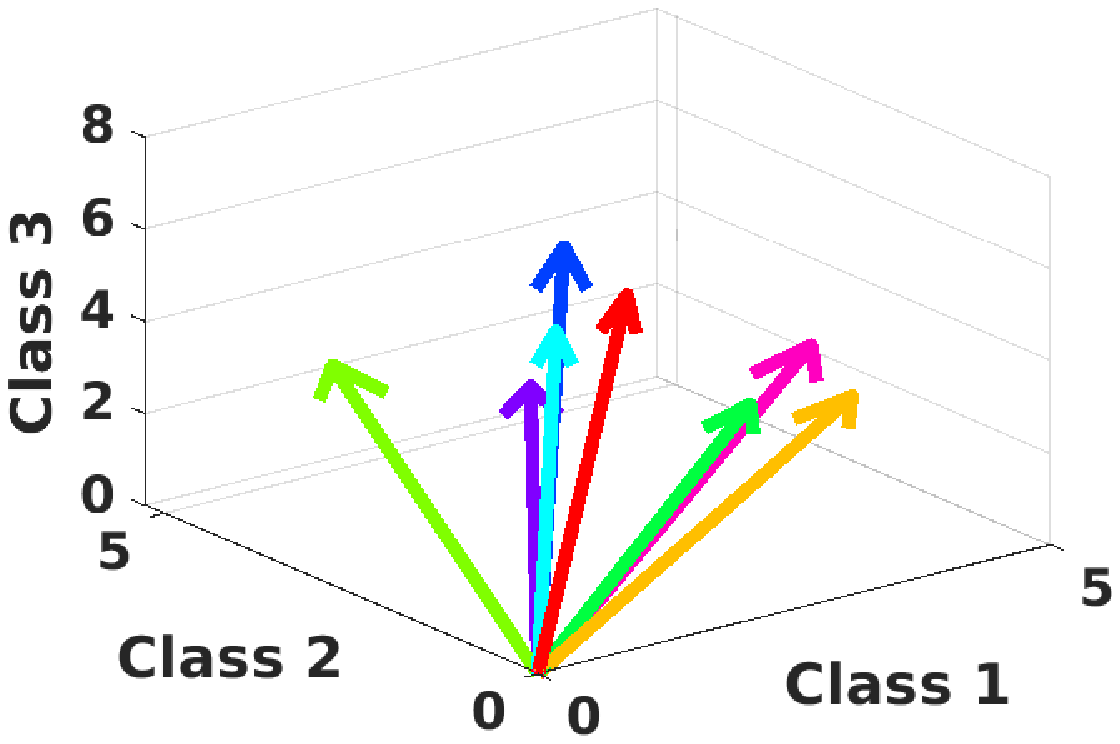}
		\caption{}	
	\end{subfigure}	
	\begin{subfigure}{0.23\textwidth}	
		\includegraphics[width=1\linewidth]{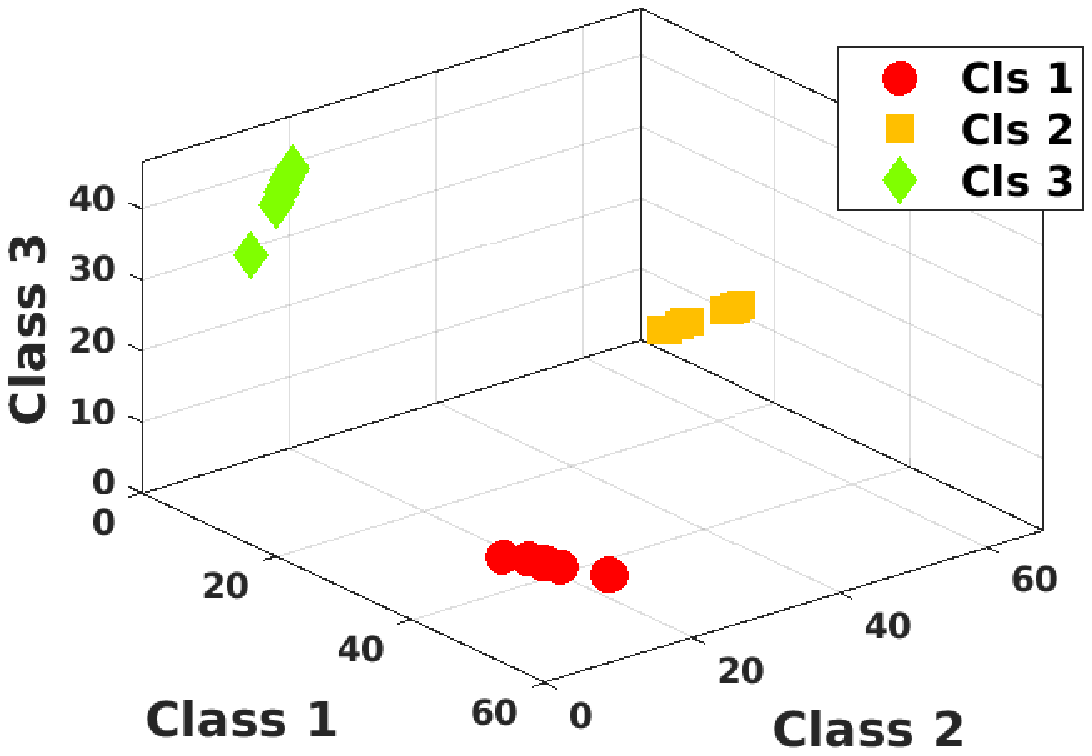}
		\caption{}
	\end{subfigure}
	\begin{subfigure}{0.23\textwidth}		
		\includegraphics[width=1\linewidth]{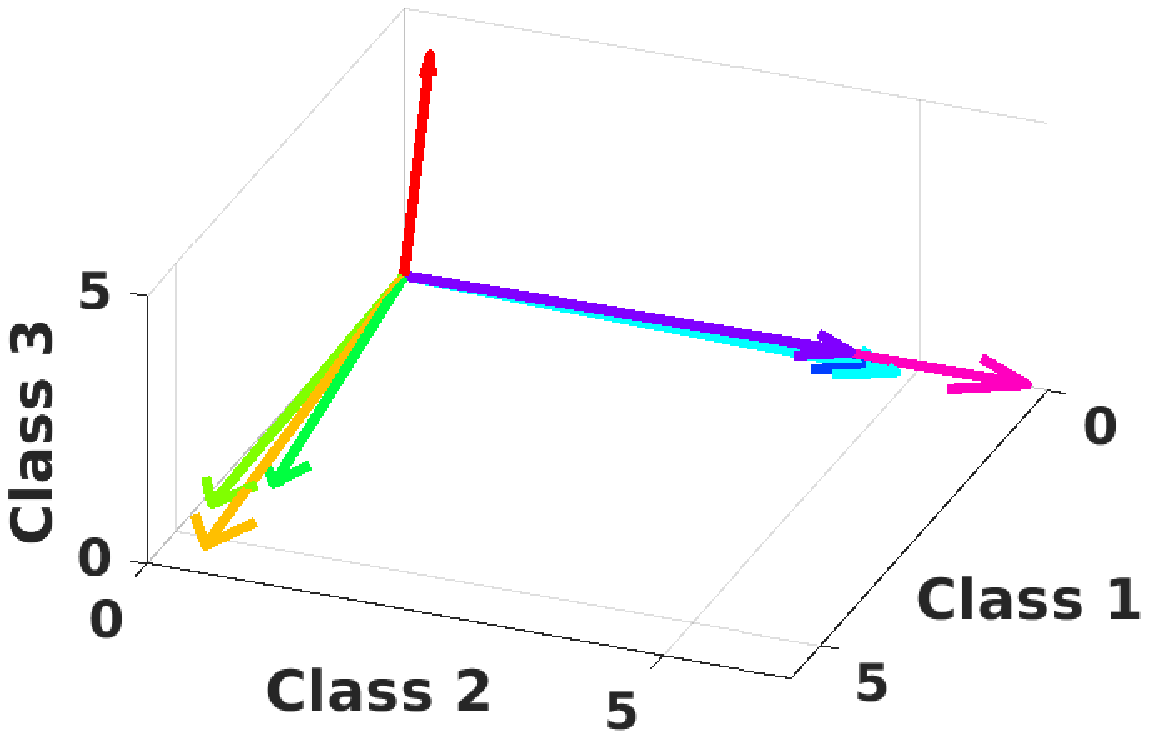}
		\caption{}
	\end{subfigure}
	\caption{When embedding vectors use all data points: (a) its projection on class-labels is coupled and (b) the embedding vectors are highly correlated in the label space. A class-based interpretable embedding: 
		(c) provides a more distinct projection on class labels and 
		(d) its dimensions can be distinguished and explained based on class labels.}
	\label{fig:interp}
\end{figure}

One of the important practical concerns regarding dimensionality reduction is the interpretation of new dimensions. 
It is common to observe in many DR methods that the embedding dimensions are constructed upon arbitrary combinations of many uncorrelated physical dimensions~\cite{tian2013interpretable,chipman2005interpretable}. Such occasions can make the interpretation of these dimensions difficult or impossible.
Such condition becomes even more severe for kernel-DR methods where the embedding dimensions are an implicit combination of data points in RKHS. 
For instance methods similar to K-PCA, the embedding vectors almost use weighted combination of all data points from all the classes. 
Hence, it would be difficult to relate any of the dimensions to any class of data (Figure~\ref{fig:interp}(a)). Furthermore, a high correlation between embedding directions can be found when considering the class-contributions in them (Figure~\ref{fig:interp}(b)).

As an improvement, sparse K-PCA~\cite{wang2016sparse} applies an $l_1$-norm 
sparsity objective to form embedding vectors from sparse combinations of training samples. 
However, these samples still belong to different classes which makes the resulting embeddings weak according to the class-based interpretation (Figure~\ref{fig:interp}).

\subsection{Motivation}
As discussed in the previous paragraphs, 
one crucial challenge for kernel-DR algorithms is the interpretation of their projection dimensions.
Based on the relation of these dimensions to the selection of data points, it is logical to focus on having each selection linked to mostly one class of data. This strategy can lead to the class-based interpretation as in Figure~\ref{fig:interp}(c)(d).

Besides, current kernel-DR methods cannot efficiently embed the multi-cluster data classes to an LD space such that the clusters could still be separated from each other. In particular, they suffer from considering the local distributions inside the classes.

Based on the current state-of-the-art, the research in kernel-DR is always distinct from feature selection on the manifold.
Although in some research, these concerns are employed in a single framework~\cite{lin2011multiple,jiang2014trace}, the DR aspect of the problem was not well investigated.
Nevertheless, in particular for discriminative tasks, these two aspects should act as each other's complements in a single framework.

\subsection{Contributions}
In this work, we propose a novel discriminative dimensionality reduction method which projects the data from an implicit RKHS space to a low-dimension vectorial space. Besides, it can join this embedding with feature selection in case of having multiple representations for the data on the manifolds.
We can summarize our contributions as follows:
\begin{itemize}
	\item We introduce the class-based interpretation concern for the kernel-DR frameworks through which the embedding dimensions can be explained according to the classes they most represent. 
	\item We show that focusing on the within-class local similarities and between-class dissimilarities can provide a more discriminative embedding.
	\item We fuse feature selection with our kernel-DR framework which leads to a more discriminative feature selection compared to the state-of-the-art.
\end{itemize}

In the rest of this paper, we provide preliminaries in Sec.~\ref{sec:pre} and discuss our discriminative kernel-DR framework in Sec.~\ref{sec:method}.
The optimization steps and the experimental results are discussed in Sec.~\ref{sec:opt} and Sec.~\ref{sec:ex} respectively.
We summarize our findings in the conclusion section.
\section{Preliminaries}\label{sec:pre}
\subsection{Notations}
We denote the matrix of training data by 
$\nY=\left[\ny_1,...,\ny_N\right] \in \mathbb{R}^{d\times N}$, and the corresponding class label matrix is given as
$\nH=[\nh_1,\dots,\nh_N] \in\{0,1\}^{\ncl \times N}$.
Each $\nh_i$ is a zero vector except in its $q$-th entry where $\nhs_{qi}=1$ if $\ny_i$ belongs to class $q$ in a $\ncl$-class setting. 
In general, for a given matrix $\MB{A}$, $\vec a_i$ denotes its $i$-th column, 
$\MB{A}(j,:)$ denotes its $j$-th row,
and $a_{ji}$ refers to the $j$-th entry in $\vec a_i$.

\subsection{Kernel-based Dimensionality Reduction}
Assume there exists an implicit non-linear mapping $\Phi(\nY)$ corresponding to the mapping of $\nY$ into an RKHS, 
which corresponds to a kernel matrix $\K(\nY,\nY)=\Phi^\top(\nY)\Phi(\nY)$.
Generally, a kernel-DR algorithm tries to obtain an embedding $\nx=\nU^\T \Phi(\ny)$ as a mapping from the features space to an LD space.
Since the dimensions of $\Phi(\ny)$ are not directly accessible in the feature space, it is common to assume embedding dimensions are constructed as 
\EQ{
\nU=\Phi(\nY)\nA,
}{\label{eq:PCA_map}}
where $\nA\in\Rnx{N}{\nd}$. 
Hence, the matrix $\nA$ projects the data from the HD feature space to a $\nd$-dimensional space,
where each embedding vector $\na_i$ is a combination of the training samples in RKHS.

Regarding the above, the K-PCA method preserves the variance of the reconstruction and to obtain embedding dimensions which are orthogonal and sorted based on their maximum variations.
To that aim, K-PCA uses the following optimization:
\begin{equation}
\begin{array}{ll}
\underset{\nA}{\min}& \| \Phi(\nY)-\Phi(\nY)\nA \nA^\T\Phi(\nY)^\T\Phi(\nY) \|_F^2\\
\mathrm{s.t.} & \nA^\T\Phi(\nY)^\T\Phi(\nY)\nA=\MB{I},
\end{array}
\label{eq:K-PCA}
\end{equation}
Although K-PCA is a powerful preprocessing algorithm to eliminate the low-variate dimensions,
it does not have any direct focus on the discrimination of the embedded data classes.
Also, each embedding vectors $\nu_i$ consists of both positive and negative contributions from all training samples which makes their interpretation difficult. 

On the other hand, the K-FDA algorithm tries to obtain an embedding $\MB{W}$ which increases the between-class covariance matrix $\MB{S}^\phi_{\MB{B}}$ while preserving the total within-class covariance matrix $\MB{S}^\phi_{\MB{W}}$ in RKHS~\cite{mika1999fisher}.
It uses the following optimization framework:
\begin{equation}
\begin{array}{l}
\underset{\MB{W}}{\max} \Tr({\MB{W}^\T \MB{S}_{\MB{B}}\MB{W}})
\qquad \mathrm{s.t.} \MB{W}^\T \MB{S}_{\MB{W}}\MB{W}=\MB{I},
\end{array}
\label{eq:K-LDA}
\end{equation}
where $\MB{W}$ has a structure analogous to Eq.~(\ref{eq:PCA_map}).
Regardless of its supervised performance, the constraint on intra-class variances can become a critical weakness when there are sub-clusters in each data class. In such cases, the constraint in Eq.~(\ref{eq:K-LDA}) cause considerable overlapping between different classes.

Our proposed framework improves the state-of-the-art in both discriminative 
kernel-DR and class-based interpretation of embedding dimensions.
\section{Interpretable Discriminative Dimensionality Reduction}\label{sec:method}
We want to obtain the embedding 
\EQ{
	\nx=\nA^\T 	\Phi(\nY)^\T\Phi(\ny)\qquad \nx\in\MBB{R}^k
}{\label{eq:AXX}}
as a projection from the original implicit RKHS to a $\nd$-dimensional explicit space which also preserves the essential characteristics of $\nY$ in the original space.
\begin{defin}
	The embedding vector $\Phi(\nY)\na_i$ is class-based interpretable 
	if we have 
	$\frac{\nH(q|\nhs_{qi}=1,:)\na_i}{\|\nH \na_i\|_1} \approx 1,$ 
	and it acts as the projection of data points on class $q$.
	\label{def:ip}
\end{defin}
In other words, $\Phi(\nY)\na_i$ can be interpreted as a projection to class $q$ if it is constructed only from that class of data.
Although Definition~\ref{def:ip} considers an ideal situation regarding the interpretability of an embedding dimension, 
we consider the value of 
\begin{equation}
{\nH(q|\nhs_{qi}=1,:)\na_i}/{\|\nH \na_i\|_1}
\label{eq:clas_i}
\end{equation}
as a measure of class-based interpretation as well.
To be more specific regarding our framework, we aim for the following objectives:

\textbf{O1:} Increasing the class-based interpretation of embedding dimensions. 

\textbf{O2:} The embedding should make the classes more separated in the LD space.

\textbf{O3:} The classes should be locally more condensed in the embedded space.

\textbf{O4:} The DR framework should also support the feature selection objective if a multiple kernel representation is provided.

%
Therefore, we formulate the following optimization scheme w.r.t. all the above objectives:
\begin{equation}
\begin{array}{ll}
\underset{\nA,\nAlf}{\min}
&\MC{J}_{Sim}+\lambda \MC{J}_{Dis}+ \mu \MC{J}_{Ip}\\
\mathrm{s.t.} & \underset{m=1}{\sum}^\nf \nalf_m=1,~ \underset{j=1}{\sum}^N \nas_{ji}=1, \forall i\\
&\nas_{ij}, \nalf_{i} \in \mathbb{R}^{+},~~\forall ij.
\end{array}
\label{eq:DR_frame}
\end{equation}
In Eq.~(\ref{eq:DR_frame}), the cost functions $\MC{J}_{Dis}$, $\MC{J}_{Ip}$, and $\MC{J}_{Sim}$ and the constraints on the optimization variables are 
designed to fulfill our research objectives \textbf{O1-O4}.
In the following sub-sections, we explain each specific term in our framework in detail and provide the rationales behind their definitions. 

\subsection{Interpretability of the Dimensions}
%
%
%
%
%
In Eq.~(\ref{eq:AXX}), each dimension $\na_i$ of the embedding is composed of a weighted selection of data points in RKHS. In K-PCA, typically all $\nas_{ji},\forall j=1,\dots,N$ have non-zero values. More specifically, for each $\na_i$, a wide range of training data from different classes are selected with large weights which weaken the interpretation of $\na_i$ regarding the class to which it could be related.
%
%

\begin{figure}[b]
	\centering
	\includegraphics[width=.61\linewidth]{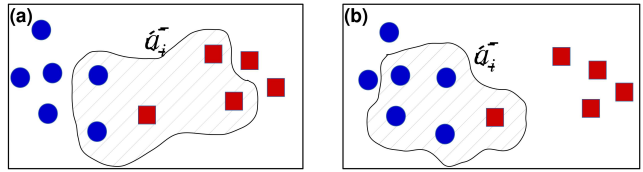}
	\caption{
		Effect of using $\MC{J}_{Ip}$ on the formation of an embedding vector $\na_i$ as the weighted combination of selected data points (inside the hatched area) in the RKHS.
		(a): Without using $\MC{J}_{Ip}$, the learned $\na_i$ cannot be assigned to either of $\{circle, square\}$ classes. 
		(b): After employing $\MC{J}_{Ip}$, the formed $\na_i$ can almost be interpreted by the \textit{circle} class.
	}
	\label{fig:lp}
\end{figure}

To make each $\na_i$ more interpretable in our framework, we propose the cost function $\mathcal{J}_{Ip}$ that its minimization enforces $\na_i$ to be constructed using similar samples in the RKHS: 
\begin{equation}
\begin{array}{l}
\mathcal{J}_{Ip}(\nY,\nA)=
\frac{1}{2}
{\underset{i=1}{\mathlarger\sum^{\nd}}}\underset{~s,t=1}{\mathlarger\sum^N}  \nas_{si}\nas_{ti} \| \Phi(\ny_s)-\Phi(\ny_t)\|_2^2 
,
\end{array}
\label{eq:j_ip}
\end{equation}
where we restrict $\nas_{ij} \ge 0, \forall {ij}$.
We call $\MC{J}_{Ip}$ as the interpretability term (Ip-term) which is an unsupervised function and independent from the value of $\nH$.
The Ip-term enforces each embedding dimension $\na_i$ to use samples in $\Phi(\nY)$ that are located in a local neighborhood of each other in RKHS (Figure~\ref{fig:lp}) by introducing a penalty term 
$\nas_{si}\nas_{ti} \| \Phi(\ny_s)-\Phi(\ny_t)\|_2^2$ on its entries.
Resulting from this term along with the non-negativity constraint on $\nA$, 
non-zero entries of $\na_i$ correspond to the neighboring points such as $(s,t)$ where their pairwise distance $\|\Phi(\ny_s)-\Phi(\ny_t)\|_2^2$ is small.
Furthermore, although Ip-term does not employ the label information, 
by assuming a smooth labeling for the data,
this regularization term constructs each $\na_i$ by contributions from more likely one particular class. 
Therefore, as a solution to our first research objective (\textbf{O1}), using Ip-term improves the class-based interpretation of $\na_i$ to relate it a sub-group of data points mostly belonging to one specific class of data (Eq.~(\ref{eq:clas_i})).
%

\subsection{Inter-class dissimilarity}
%
Regarding our second objective (\textbf{O2}), we focus on increasing the inter-class dissimilarities in the LD space which makes the embedded classes more distinct.
To that aim, we define the loss term $\mathcal{J}_{Dis}$ as
\begin{equation}
\begin{array}{l}
\mathcal{J}_{Dis}(\nY,\nH,\nA)=\\ 
\Tr(\overline{\nH}^\top\nH\Phi(\nY)^\T\Phi(\nY)\nA\nA^\T\Phi(\nY)^\T\Phi(\nY)),
\end{array}
\label{eq:J_dis}
\end{equation}
where $\overline{\nH}$ is the logical complement of $\nH$.
Throughout simple algebraic operations, we can show that Eq.~(\ref{eq:J_dis}) is the reformulation of
\EQ
{
	\underset{i}{\Bsum}\underset{j|\nh_j \neq \nh_i}{\Bsum}
	\langle \nA^\T\Phi(\nY)^\T\Phi(\ny_i), \nA^\T\Phi(\nY)^\T\Phi(\ny_j)\rangle.
}
{\label{eq:J_dis2}}
Hence, minimizing $\MC{J}_{Dis}$ motivates the global separation of the classes in the embedded space by reducing the similarity between their projected vectors 
$\nA^\T\Phi(\nY)^\T\Phi(\ny)$.
%
\subsection{Intra-class similarities}
%

Even though the introduced cost term $J_{Dis}$ helps the embedded classes to obtain more distance from each other, 
it still does not consider the intra-class similarities which concerns our third objective (\textbf{O3}).
It is important to note that we want to make the projected vectors $\nx_i$ of each class more similar to each other, 
while still preserving the local structure of the class respecting the possible sub-classes. 
This characteristic works against the drawback of K-FDA when facing distinct sub-classes as pointed out by \cite{liu2007pairwise}. 

To address the above concern, we proposed the following cost function
\EQ{
	\MC{J}_{Sim}=\sum_{i=1}^N (\nH(q|\nhs_{qi}=1,:) \nA \nA^\T \Phi(\nY)^\T \Phi(\ny_i)-1)^2,
}{\label{eq:sim}}
in which $q$ is the class to which $\ny_i$ belongs.
Furthermore, based on Eq.~(\ref{eq:DR_frame}), we apply an affine constraint on columns of $\nA$ as $\| \na_s \|_1=1, \forall s=1,\dots,N$.
By combining Eq.~(\ref{eq:sim}) with $\nx_i$ from Eq.~(\ref{eq:AXX}) we have
\begin{equation}
\MC{J}_{Sim}=\sum_{i=1}^N (\nH(q|\nhs_{qi}=1,:) \nA \nx_i-1)^2,
\end{equation}
which applies constraints on columns of $\nA$ corresponding to large entries of $\nx_i$.
Specifically, those constraints aim the entries which are related to the data points which have the same label as $\ny_i$.
For instance, if $\nxs_{si}$ has a relatively large value, minimizing $\MC{J}_{Sim}$ optimizes the entries $\nas_{js}$ where $\nh_j=\nh_i$. 
Besides, the applied $l_1$-norm sparsity constraint $\|\na_s\|_1=1$ enforces some entries in $\na_s$ to shrink near to zero.
Therefore, it is simple to conclude that these entries would mostly include $\nas_{js}$ where $\nh_j\neq\nh_i$.

On the other hand, $\nxs_{si}=\sum_{t=1}^N \nas_{ts}\Phi(\ny_t)^\T\Phi(\ny_i)$. 
Hence, Having the $l_1$-norm of $\na_s$ restricted along with its non-negativity constraint naturally motivates the optimization process to assign large values to entries $\nas_{ts}$ corresponding to data points $\ny_t$ with large $\Phi(\ny_t)^\T\Phi(\ny_i)$.
In other words, $\na_s$ selects the nearby data points of $\ny_i$ as its most similar neighbors. 
Combining this finding with our first conclusion about the effect of Eq.~(\ref{eq:sim}), along with the localization role of $\MC{J}_{Ip}$,  
minimizing $\MC{J}_{Sim}$ helps each data point $\ny_i$ to be encoded in particular by its nearby embedding vectors $\na_s$, 
which are also constructed mostly by the same-class of samples in the vicinity of $\ny_i$ (\textbf{O1}).    
Consequently, the data points from each local sub-class are embedded by similar sets of columns in $\nA$
and are mapped into a local neighborhood in the LD space.
In other words, This embedding increases the intra-class similarities for the projected columns in $\nX=[\nx_1,\dots,\nx_N]$. 
\subsection{Feature Selection on the Manifold}
It is a feasible assumption for any structured and non-structure $\nY$ to 
have $\nf$ different kernel representations available \cite{bach2004multiple}, 
such that each $\K_m(\nY,\nY), \forall m=1,\dots,\nf$, maps the $m$-th dimension of the original data into an RKHS or is derived from the $m$-th descriptor (e.g., for images).
Given the above, we can assume 
\begin{equation}
\begin{array}{l}
\Phi(\ny)=
[ \phi_1^\top(\ny),\dots,
 \phi_{d}^\top(\ny)]^\top,
\end{array}
\end{equation}
where each 

$$\mathcal{S}:\MBB{R}^D \rightarrow \MBB{R}^d, ~~d << D$$

represents an implicit mapping from the original space to a subspace of the RKHS,
such that $\K_m(\ny_t,\ny_s)=\phi_m^\top(\ny_t)\phi_m(\ny_s)$.
Therefore, we can consider a diagonal matrix $\nALF \in \MBB{R}^{d\times d}$ which  
provides scaling of the RKHS by 
\begin{equation}
\small
\begin{array}{l}
\phih(\ny)=\nALF \Phi(\nY)=
[\sqrt{\nalf_1} \phi_1^\top(\ny),\cdots,\sqrt{\nalf_\nf} \phi_{\nf}^\top(\ny)]^\top,
\end{array}
\label{eq:mk}
\end{equation}
where $\nAlf$ is the vector of combination weights derived from diagonal entries of $\nALF$. 
We can compute the weighted kernel matrix $\hat{\K}$ corresponding to $\phih(\nY)$ as
\begin{equation}
\begin{array}{l}
\hat{\K}(\ny_t,\ny_s)=\sum_{m=1}^{\nf} \nalf_m \K_m(\ny_t,\ny_s).
\end{array}
\label{eq:K_alf}
\end{equation}
Additionally, we apply  
a non-negativity constraint on entries of $\nAlf$ as $\nalf_i \ge 0$ to make the resulted kernel weights interpreted as the relative importance of each kernel in the weighted representation $\phih(\nY)$~\cite{gonen2011multiple}.
Consequently, we can obtain a feature selection profile by sorting entries of $\nAlf$ based on their magnitude.
For the ease of reading, in the rest of the paper, we denote $\hat{\K}(\nY,\nY)$ and ${\K_i}(\nY,\nY)$ by $\hat{\K}$ and ${\K_i}$ respectively.

Substituting $\Phi(\nY)$ by $\phih(\nY)$ in the definitions of $\MC{J}_{Dis}$, $\MC{J}_{Ip}$, and $\MC{J}_{Sim}$ reformulates them also as a function of $\nALF$.
Therefore, minimizing those terms also optimizes the value of $\nALF$ regarding their specific purposes.
Furthermore, we apply an $l_1$-norm restriction on the value of $\nALF$ as the affine constraint $\sum_{m=1}^\nf \nalf_m=1$.
This constraint prevents $\nAlf$ from becoming a vector of zeros as the trivial solution 
and additionally results in a sparse feature selection to reduce the redundancies between different kernel representations~\cite{rakotomamonjy2008simplemkl}.
We can claim that by using $\phih(\nY)$ in each of the defined terms, the resulted feature selection also complies with those specific characteristics.     
%
%
%
In the next section, we discuss the optimization scheme of Eq.~(\ref{eq:DR_frame}).
\subsection{Optimization Scheme}\label{sec:opt}
The cost function $\MC{J}_{Sim}$ is non-convex which makes the objective function of Eq.~(\ref{eq:DR_frame}) non-convex as well.
Hence, we define a variable matrix $\MB{S}$ and relax Eq.~(\ref{eq:DR_frame}) to the following optimization problem 
\begin{equation}
\begin{array}{ll}
\underset{\nA,\nAlf,\MB{S},\nX}{\min}
&
\sum_{i=1}^N (\nH(q|\nhs_{qi}=1,:) \vec s_i-1)^2\\
&+\lambda \Tr(\nA^\T \hat{\K}\overline{\nH}^\top\nH\hat{\K}\nA)
+ \mu \Tr(\nA^\T\tilde{\K}\nA)\\
&+\tau\|\MB{S}-\nA\nX\|_F^2
+\zeta\|\nX-\nA^\T\K\|_F^2\\
\mathrm{s.t.} & \underset{m=1}{\sum}^\nf \nalf_m=1,~ \underset{j=1}{\sum}^N \nas_{ji}=1, \forall i\\
&\nas_{ij}, \nalf_{i} \in \mathbb{R}^{+},~~\forall ij
,
\end{array}
\label{eq:relaxed}
\end{equation}
in which 
$\tilde{\K}={diag(\hat{\K}\vec{\MB{1}}})-\hat{\K}$, and the operator $diag(.)$ creates a diagonal matrix from its vector argument.
The constants $\lambda, \mu$ are the control parameters for the role of introduced loss terms in the optimization scheme, 
and the constants $\tau, \zeta$ should be large enough to make sure the slack variables $\MB{S}, \nX$ have appropriate values.
The second and third parts of the objective in Eq.~ (\ref{eq:relaxed}) are reformulations of $\MC{J}_{Dis}$ and $\MC{J}_{Ip}$,
which can be obtained by using the \textit{kernel trick} and the Laplacian matrix \cite{von2007tutorial}. 
We initialize the embedding matrix $\nA$ using random entries and adjust its columns to have unit $l_1$-norm.
Then, we optimize $\nX, \MB{S}, \nA$, and $\nAlf$ alternatively based on the following steps.

(1) Fix $\MB{S}, \nA$, and $\nAlf$ and update $\MB{\nX}$ as:
\EQ{
	\nX^*=\nA^\T \hat{\K}.
}{\label{eq:upx}}

(2) Fix $\nX, \nA$, and $\nAlf$ and update $\MB{S}$:
\begin{equation}
\begin{array}{lll}
\vec{s_i}^*=\underset{\vec{s_i}}{\argmin}~
\vec{s_i}^\T(\vec{u_i}^\T \vec{u_i}+\MB{I})\vec{s_i}-2(\vec{u_i}+\nx_i^\T\nA^\T)\vec{s_i},
\end{array}
\end{equation}
where $\vec{u_i}=\nH(q|\nhs_{qi}=1,:)$. This unconstrained quadratic programming has the closed-form solution
\begin{equation}
	\vec{s_i}^*=(\vec{u_i}^\T \vec{u_i}+\MB{I})^{-1}(\vec{u_i}+\nx_i^\T\nA^\T)^\T.
\end{equation}

(3) Fix $\nX, \MB{S}$, and $\nAlf$ and update $\MB{\nA}$ as:
\begin{equation}
\begin{array}{lll}
\nA^*=&\underset{\nA}{\argmin}
&\lambda \Tr(\nA^\T \hat{\K}\overline{\nH}^\top\nH\hat{\K}\nA)
+ \mu \Tr(\nA^\T\tilde{\K}\nA)\\
&&+\tau\|\MB{S}-\nA\nX\|_F^2
+\zeta\|\nX-\nA^\T\K\|_F^2\\
&\mathrm{s.t.} 
& ~\nA^\T \1=\1,~\nas_{ij}\in \MBB{R} ^{+},  \forall ij.\\
\end{array}
\label{eq:up_A}
\end{equation}
Calling the objective of Eq.~(\ref{eq:up_A}) $\MC{J}_{\nA}$, it is possible to show that $\MC{J}_{\nA}$ consists of 
convex parts and its gradient w.r.t. $\nA$ can be computed as:
\EQ{
	\nabla_\nA \MC{J}_{\nA}=\Omega \nA + \Psi,
}{} 
where $(\Omega, \Psi)$ can be obtained by simple algebraic operations.
Therefore, we use the direction method of multipliers (ADMM) \cite{boyd2011distributed} by defining the Lagrangian formulation for Eq.~(\ref{eq:up_A}):
\EQ{
	\begin{array}{l}
		\MC{L}_{\rho}(\nA,\nA_+,\Delta,\delta)\\
		=\JA
		+\frac{\rho}{2}\|\nA-\nA_{+}\|_2^2
		+\frac{\rho}{2}\|\nA^\T\1-\1\|_2^2\\
		+tr(\Delta^\T(\nA-\nA_{+}))
		+\delta^\T(\nA^\T\1-\1),
	\end{array}
}{}
and following these steps:
\begin{equation}
\begin{cases}
\nA^{(t+1)}=\underset{\nA}{\argmin} \MC{L}_{\rho}(\nA,\nA_+,\Delta,\delta),\\
\nA_{+}^{(t+1)}=\text{max}(\nA^{(t+1)}+\frac{1}{\rho}\Delta^{(t)},0),\\
\Delta^{(t+1)}=\Delta^{(t)}+\rho(\nA^{(t+1)}\1-\1),\\
\delta^{(t+1)}=\delta^{(t)}+\rho(\nA^{(t+1)}-\nA_{+}^{(t+1)}),\\
\end{cases}
\label{eq:ADMM}
\end{equation}
In Eq.~(\ref{eq:ADMM}), $\nA_+$ is an axillary matrix related to the non-negativity constraint, $\Delta \in \MBB{R}^{N\times N}$ and $\vec{\delta} \in \MBB{R}^{N}$ are the Lagrangian multipliers, and $\rho \in \MBB{R}^+$ is the penalty parameter. 
We update the matrix $\nA^{(t+1)}$ based on its closed-form solution derived from having $\nabla_\nA \MC{L}_{\rho}=0$. 

(4) Fix $\nX, \MB{S}$ and $\nA$ and update $\MB{\nAlf}$:
By combining Eq.~(\ref{eq:K_alf}) and Eq.~(\ref{eq:relaxed}) and removing the constant terms, $\nAlf$ can be updated by the following quadratic programming (QP)
\begin{equation}
\begin{array}{lll}
\nAlf^*=&\underset{\nAlf}{\argmin}
&\frac{1}{2}\nAlf^\T \MB{Q} \nAlf+ \vec{v}^\T \nAlf,\\
&\mathrm{s.t.} 
&\nAlf^\T \1=1,~\nalf_{i}\in \MBB{R} ^{+}, \forall i.
\end{array}
\label{eq:QP}
\end{equation}
In this formulation, $\forall ij=1,\dots,\nf$:
\EQ{
\MB{Q}_{ij}=
\lambda \Tr(\nA^\T \hat{\K}_i\overline{\nH}^\top\nH\hat{\K}_j\nA)
+ \zeta \Tr(\hat{\K}_i \nA^\T \nA \hat{\K}_j),
}{} 
and 
\EQ{
v_i=\mu \Tr(\nA^\T\tilde{\K}_i\nA)-2\Tr(\nX^\T\nA^\T\hat{\K}_i).
}{} 
The optimization problem of Eq.~(\ref{eq:QP}) is an instance of constraint quadratic programming and can be efficiently solved by QP solvers such as CGAL\cite{gartner2000efficient} or MOSEK~\cite{mosek2015mosek}.

As a result, in each iteration of the main optimization loop, we compute the closed-form solution of $\nX, \MB{S}$ and update $\nA, \nAlf$ rapidly using the ADMM and QP solvers respectively.
The precise implementation of our kernel-DR framework is available on the online repository\footnote{https://github.com/bab-git/} 


%
\subsection{Time Complexity of the Algorithm}
In the training phase, we update $\nA, \MB{S}, \nX$, and $\nAlf$ alternatively. 
For each iteration of the algorithm, the variables $\{\nA, \MB{S}, \nX,\nAlf\}$ are updated with the time complexities of 
$\MC{O}(\MB{M}(\nd^3+\nd^2N+\nd N^2))$, $\MC{O}(N(N^3+N))$, $\MC{O}(\nd N)$, 
and $\MC{O}(\nf^2(\nd\ncl+\nd N+\nd^2)+\nf(\nd^2+\nd N)+\nf^2L)$ respectively,
where $\MB{M}$ is the number of iterations which takes for the ADMM algorithm to update $\nA$,
and $\MC{O}(\nf^2L)$ is the time complexity of the QP for updating $\nAlf$.
In practice, values of $\nd, \ncl$, and $\nf$ are much smaller than $N$. 
Hence, the computationally expensive part of the algorithm is due to 
computing the inverse of 
$(\vec{u_i}^\T \vec{u_i}+\MB{I})^{-1}$ to update each column of $\MB{S}$.
However, this particular computation is independent of update rules in the iterations, and we conduct it only once in the initialization phase of the algorithm, which considerably accelerates the convergence speed. 
\section{Experiments}\label{sec:ex}
In this section, we implement our proposed I-KDR algorithm on real-world datasets to analyze its DR and feature selection performance.
For all the datasets we compute the kernels based on the Gaussian kernel function
\begin{equation}
\K(\ny_i,\ny_j)=exp(-{\|\ny_i-\ny_j\|_2^2}/\delta),
\label{eq:gausker}
\end{equation}
in which
$\delta$ denotes the average of $\|\ny_i-\ny_j\|_2$ for all training samples.

\subsection{Datasets}
We implement our DR algorithm on real-world benchmark datasets including
Yale face recognition$\footnote{http://cvc.yale.edu/projects/yalefaces/yalefaces.html}$,
$\{$Sonar, Dbworld$\}$ from the UCI repository$\footnote{http://archive.ics.uci.edu/ml/datasets.html}$,
XM2VTS50 image dataset~\cite{messer1999xm2vtsdb},
the text datasets 20newsgroups7$\footnote{http://qwone.com/~jason/20Newsgroups/}$,
and $\{$Colon, Gli85, Central-Nervous-System (CNS)$\}$ from the feature selection repository$\footnote{http://featureselection.asu.edu/datasets.php}$.
For the 20newsgroups7 dataset, we choose the large topic \textit{comp}, and for Colon and Gli35 datasets we use the first two classes. 
The characteristics of the datasets are reported in Table~\ref{tab:data}

We evaluate the performance of the algorithms based on the average classification accuracy with 10-fold cross-validation (CV), and we use the $1$-nearest neighbor method (1-NN) to predict the label of test data based on $\nX$ of the training set.
Moreover, the parameters $\lambda$ and $\mu$ are tuned based on conducting CV on the training sets.
The same policy is applied to the selected baseline algorithms.

\begin{table}[!t]
	\centering
	\footnotesize
	\caption{Selected datasets.
		\small\{\textbf{Dim}: $\#$dimensions, \textbf{Cls}: $\#$classes, \textbf{Num}: $\#$data samples\}.
	}
	\begin{tabular}{l|c|c|c||l|c|c|c  } %
		\hline	
		Dataset &Num & Dim & Cls&Dataset&Num & Dim & Cls\\
		\hline
		\hline
		Yale&165&1024&15&Gli85&85&22283&2\\
		Sonar&208&60&2&CNS &60&7129&2\\
		Colon&62&2000&2&Dbwork&64&4702&2\\
		20NG&4852&28299&4&XM2VTS50&1180&1024&20\\
		\hline
	\end{tabular} 			
	\label{tab:data} 
\end{table}

\subsection{Dimensionality Reduction}
In this section, we only evaluate the dimensionality reduction performance of our I-KDR in a single-kernel scenario, meaning that
we use $\K$ in Eq.~(\ref{eq:relaxed}) instead of $\hat{\K}$, and $\nAlf$ is not involved in the framework.
As baseline kernel-DR methods, we choose the supervised algorithm
K-FDA,
LDR~\cite{suzuki2013sufficient},
SDR~\cite{orlitsky2005supervised},
KDR~\cite{fukumizu2004dimensionality}, and unsupervised DR algorithms
JSE~\cite{lee2013type},
SKPCA~\cite{das2018sparse}, 
and KEDR~\cite{alvarez2017kernel}.
The classification results are reported in Table~\ref{tab:DR}. 

We can observe that I-KDR obtains better performance than baselines on almost all selected datasets.
For the Colon dataset, I-KDR obtained $8.26\%$ higher accuracy than the best approach.
We can conclude that our designed elements of Eq.~(\ref{eq:DR_frame}) results in better discriminative projections than other baselines.
Regarding other algorithm, the supervised methods (e.g., LDR and SDR) generally outperform the unsupervised ones which is due to their advantage of using the supervised information in the trainings.
For Sonar and Dbwork datasets, LDR almost achieved a performance comparative to I-KDR.
\begin{table}[!b]
	\centering
	\footnotesize
	\caption{Classification accuracies ($\%$) on the selected datasets.}	
	\begin{tabular}{l|c|c|c|c|c|c|c|c} %
		\hline	
		Dataset &I-KDR & LDR & SDR&KDR&K-FDA&JSE&KEDR&SKPCA\\
		\hline
		\hline
		Yale&\textbf{79.43} &72.80 &71.13 &69.50 &67.88 &66.23 &64.61&60.75\\
		Sonar&87.01 &86.79 &84.59 &85.92 &83.45 &81.11 &82.44 &71.26 \\
		Colon&\textbf{83.37} &75.09 &74.03 &73.19 &72.05 &70.81 &70.00 &68.12\\
		20NG&\textbf{85.74} &80.76 &79.62 &80.18 &78.99 &77.82 &76.82 &72.73\\					
		Gli85&\textbf{76.45} &72.15 &70.66 &69.26 &67.50 &65.79 &66.68 &61.38\\
		CNS &\textbf{72.96} &68.77 &67.09 &65.84 &64.61 &63.21 &63.96 &58.93\\
		Dbwork&88.24 &87.67 &86.28 &84.90 &83.27 &81.74 &80.40 &77.32\\
		XM2VTS50&\textbf{95.51} &92.67 &91.62 &92.17 &90.88 &89.52 &88.55 &84.86\\
		\hline
	\end{tabular} 			
	\\
	\scriptsize{The best result (\textbf{bold}) is according to a two-valued t-test at a $5\%$ significance level.}
	\label{tab:DR} 
\end{table}

\begin{figure}[t]
	\centering
	\begin{subfigure}{0.32\textwidth}		
		\includegraphics[width=1\textwidth]{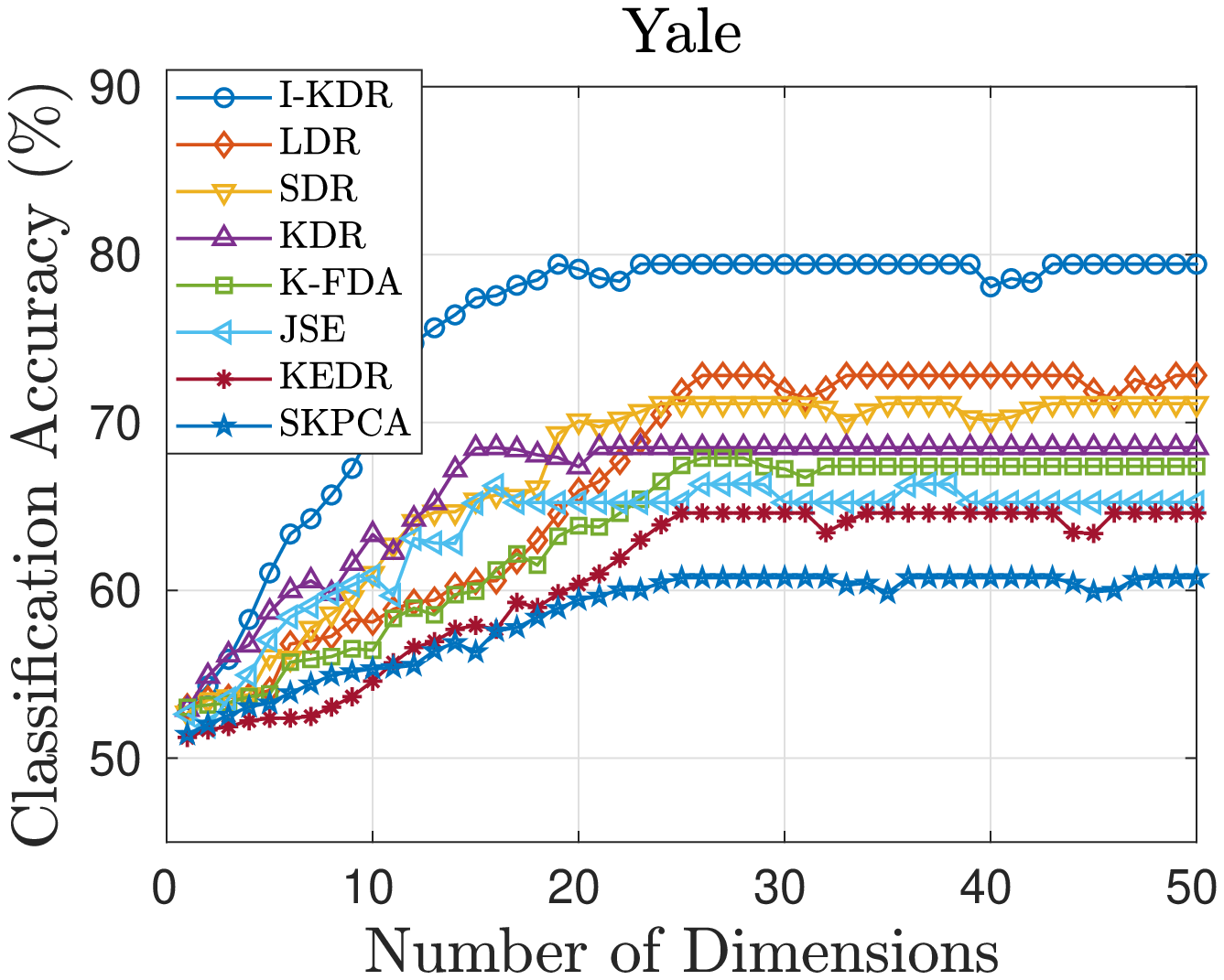}
	\end{subfigure}
	\begin{subfigure}{0.32\textwidth}	
		\includegraphics[width=1\textwidth]{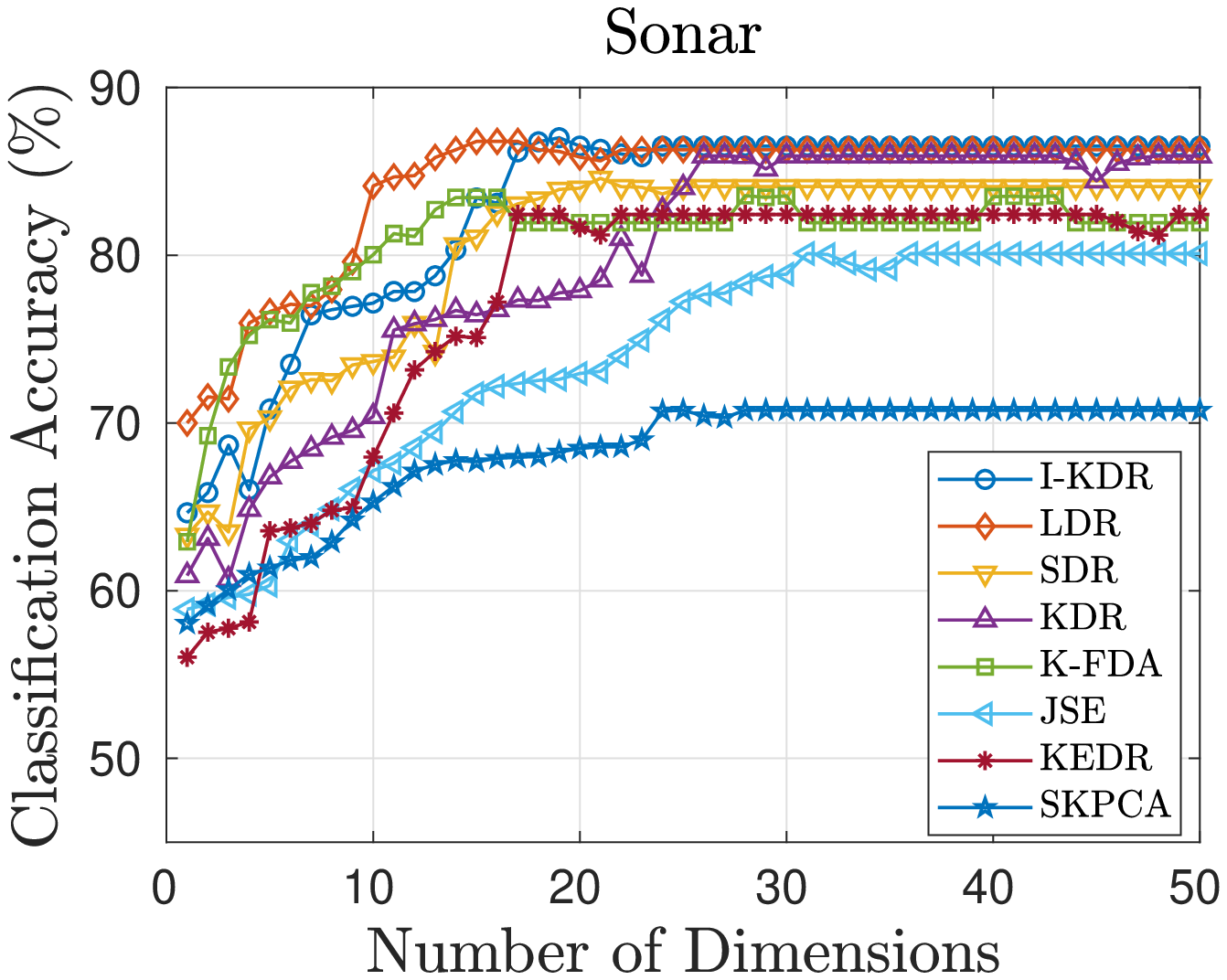}
	\end{subfigure}
	\begin{subfigure}{0.32\textwidth}
		\includegraphics[width=1\textwidth]{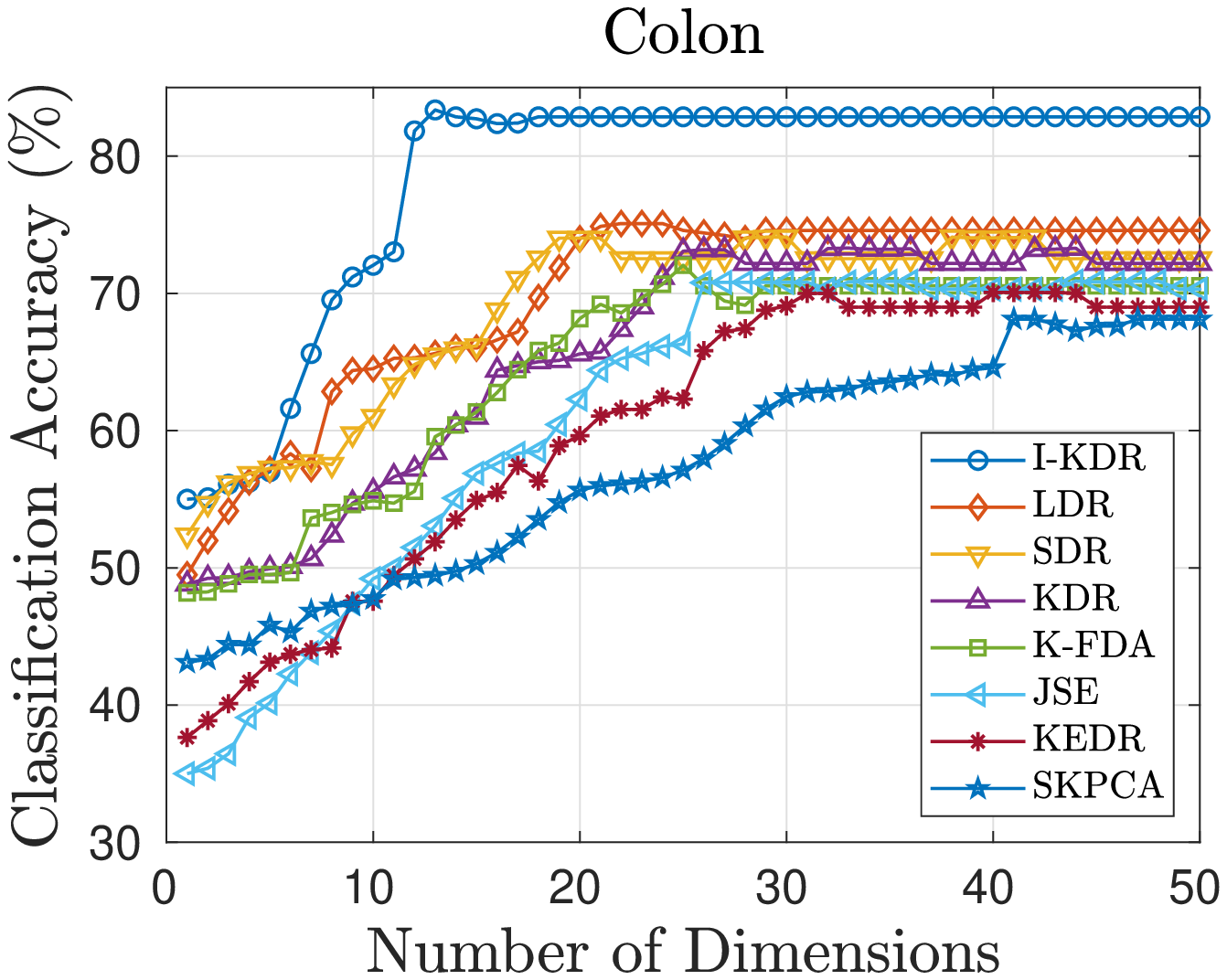}			
	\end{subfigure}
	\vfill		
	\begin{subfigure}{0.32\textwidth}
		\includegraphics[width=1\textwidth]{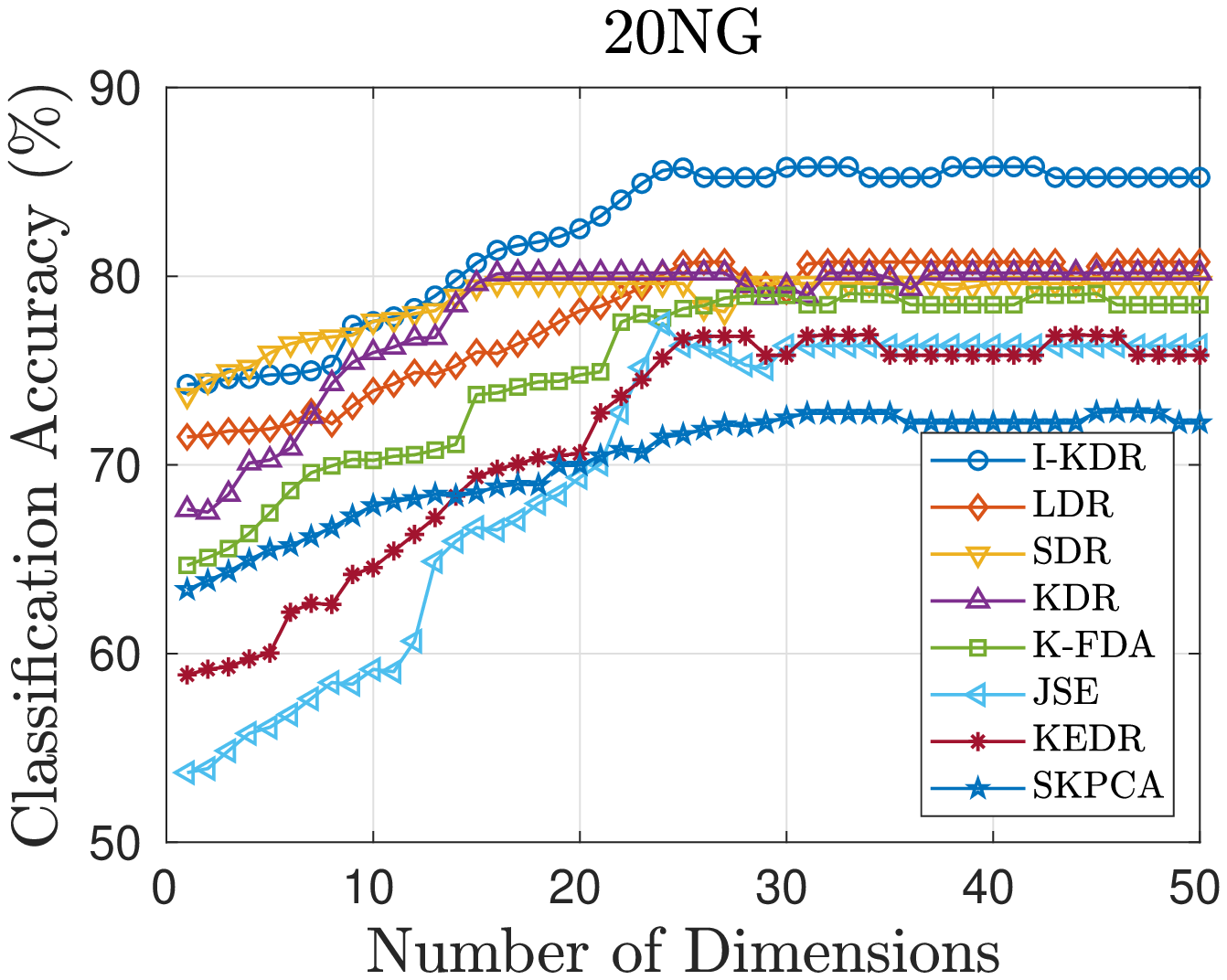}
	\end{subfigure}		
	\begin{subfigure}{0.32\textwidth}
		\includegraphics[width=1\textwidth]{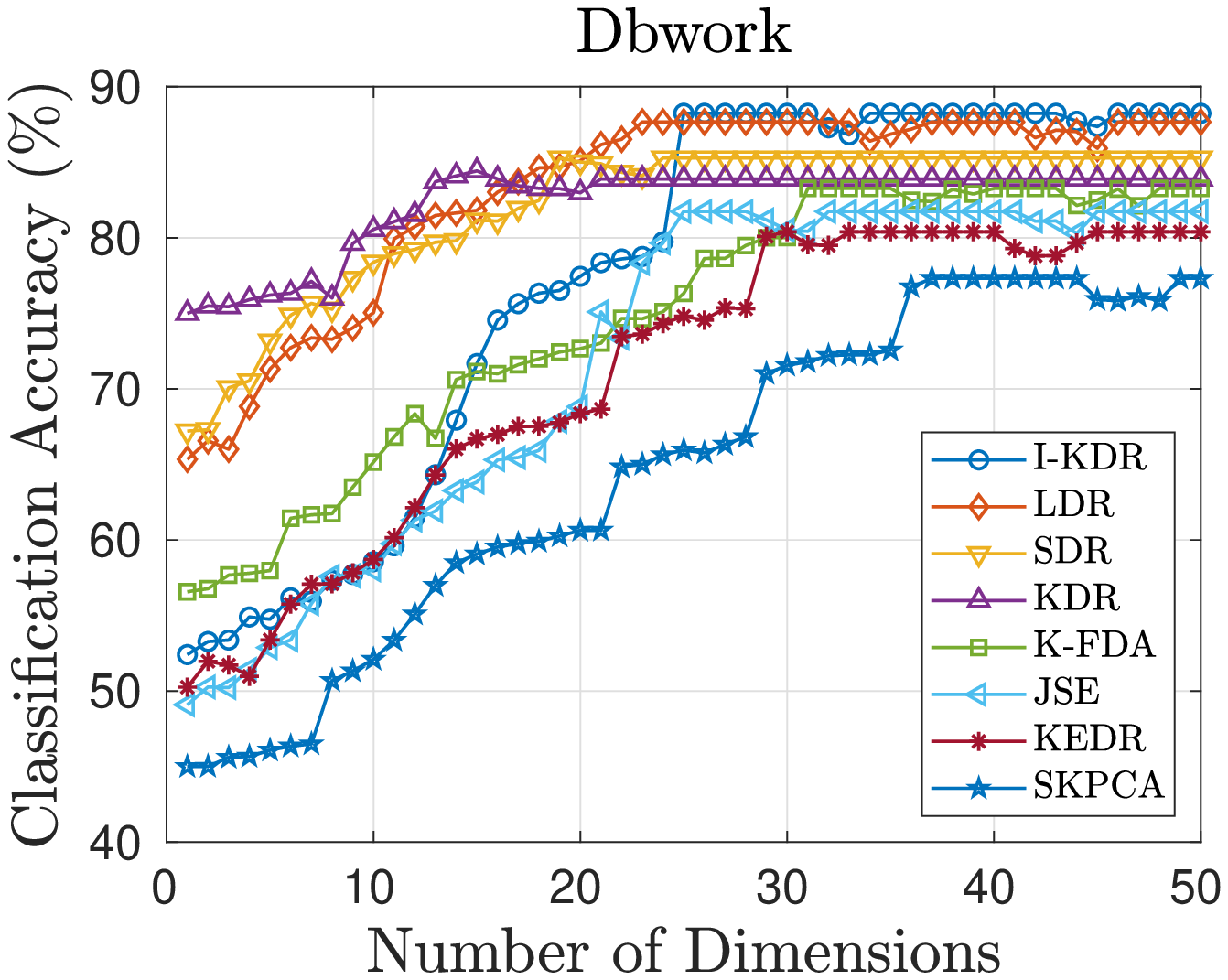}
	\end{subfigure}		
	\begin{subfigure}{0.32\textwidth}
		\includegraphics[width=1\textwidth]{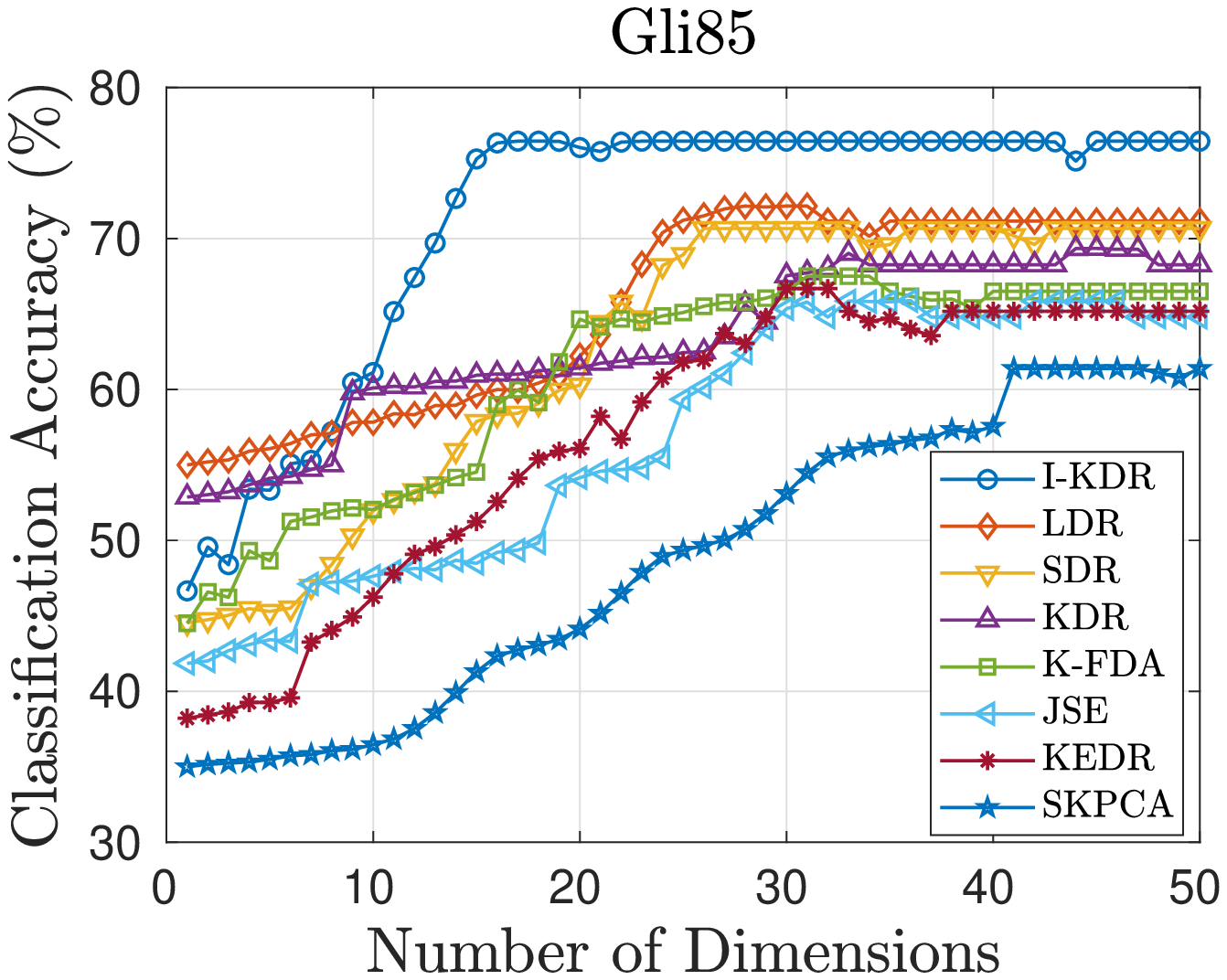}
	\end{subfigure}			
	\caption{Classification accuracy ($\%$) of the baselines respect to the number of selected dimensions for the datasets Yale, Sonar, Colon, 20NG, Dbwork, and Gli85.}
	\label{fig:dim}
\end{figure}

In Figure~\ref{fig:dim}, we compare the classification accuracy of the baselines for different numbers of selected dimensions. 
Based on the accuracy curves, I-KDR shows a distinct performance compared to other methods for the datasets Yale, Colon, and Gli85. Especially for the high-dimensional datasets Colon and Gli85, our DR algorithm achieves the peak of its performance for a smaller number of selected dimensions in comparison. For Sonar and Dbwork, I-KDR algorithm shows a competitive performance to the best baseline (LDR algorithm).
Considering the classification accuracies for Yale dataset in Figure~\ref{fig:dim}, I-KDR's curve reaches the peak accuracy of each baseline while selecting fewer dimensions for the embeddings.
Regarding the baseline DR algorithms, the supervised methods generally outperform the unsupervised algorithms in both the accuracy and number of selected dimension. This finding also complies with the reported information in Table~\ref{tab:DR}.
Therefore, applying constraints regarding the interpretability of the DR model in I-KDR does not sacrifice its discriminative performance.

\subsection{Interpretation of the Embedding Dimension}
To evaluate the effect of $\MC{J}_{Ip}$ in Eq.~(\ref{eq:DR_frame}), 
we use the $Ip$ measure defined as 
$
Ip=\frac{1}{\nd}\sum_{i=1}^\nd{(\max_{q}~\nH(q,:)\na_i)}/\|\nH \na_i\|_1.
$
The $Ip$ value considers the interpretability of each $\na_i$ based on the data points from which it is constructed.
Assuming there exists considerable similarities between the class members in RKHS, a highly interpretable embedding dimension would be formed by contributions taken from mostly one class of data. In such a case, the value of $Ip$ should grow towards $1$.
Table~\ref{tab:Ip} reports the value of this measure for those experiments in Table~\ref{tab:DR} where computing $Ip$ is possible.
Based on the results, I-KDR obtained the most interpretable embeddings among other baselines,
K-FDA has the weakest $Ip$ performance while SKPCA and KDR are jointly the runner up methods in this Table. 
Regardless of the interpretation-effective sparsity term of SKPCA, its unsupervised model allows cross-class contributions to happen in the formation of the columns of $\nA$.
From another point of view,
for Yale and CNS datasets, I-KDR has smaller $Ip$ values compared to XM2VTS and 20NG datasets for instance.
This difference happened due to substantial overlapping of the classes in the first group of datasets.   

\begin{table}[!b]
	\centering
	\footnotesize
	\caption{Comparison of the $Ip$ measure between the baselines.}	
	\begin{tabular}{l|c|c|c|c|c} %
		\hline	
		Dataset &I-KDR & SKPCA&KDR&SDR&K-FDA\\
		\hline
		\hline
		Yale&\textbf{0.80}&0.64 &0.61 &0.58 &0.55\\
		Sonar&\textbf{0.88} &0.64 &0.66 &0.63 &0.57\\
		Colon&\textbf{0.91}&0.72 &0.69 &0.66 &0.63\\
		20NG&\textbf{0.94} &0.75 &0.77 &0.73 &0.64\\				
		Gli85&\textbf{0.84}&0.69 &0.64 &0.59 &0.57\\
		CNS &\textbf{0.83}&0.66 &0.67 &0.66 &0.63\\
		Dbwork&\textbf{0.86}&0.73 &0.77 &0.70 &0.61\\
		XM2VTS50&\textbf{0.96}&0.82 &0.86 &0.79 &0.60\\
		\hline
	\end{tabular} 			
	\\
	\scriptsize{The best result (\textbf{bold}) is according to a two-valued t-test at a $5\%$ significance level.}
	\label{tab:Ip} 
\end{table}

\begin{figure}[!t]
	\centering		
	\begin{subfigure}{0.24\textwidth}	
		\includegraphics[width=1\textwidth]{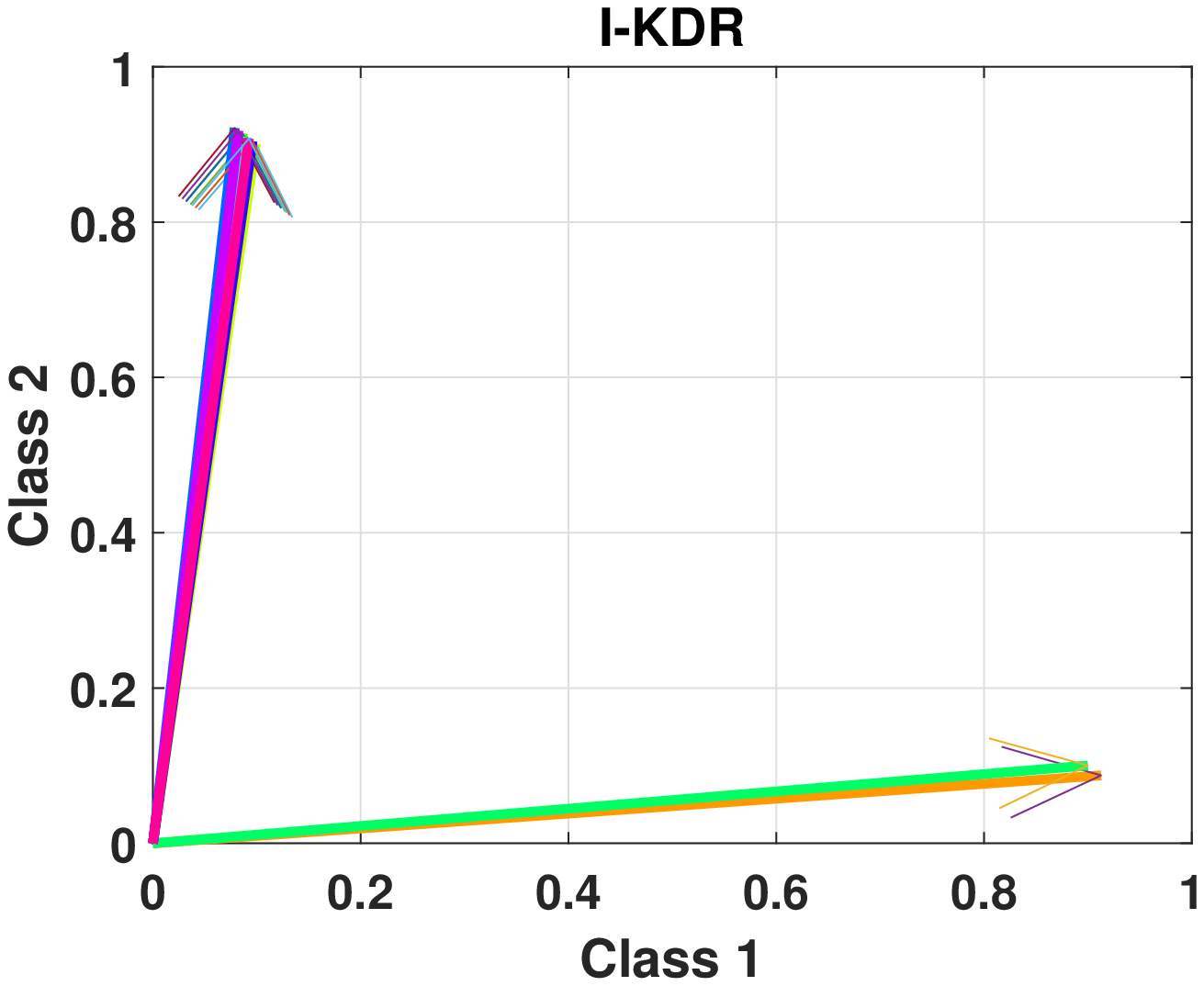}
	\end{subfigure}	
	\begin{subfigure}{0.24\textwidth}	
		\includegraphics[width=1\textwidth]{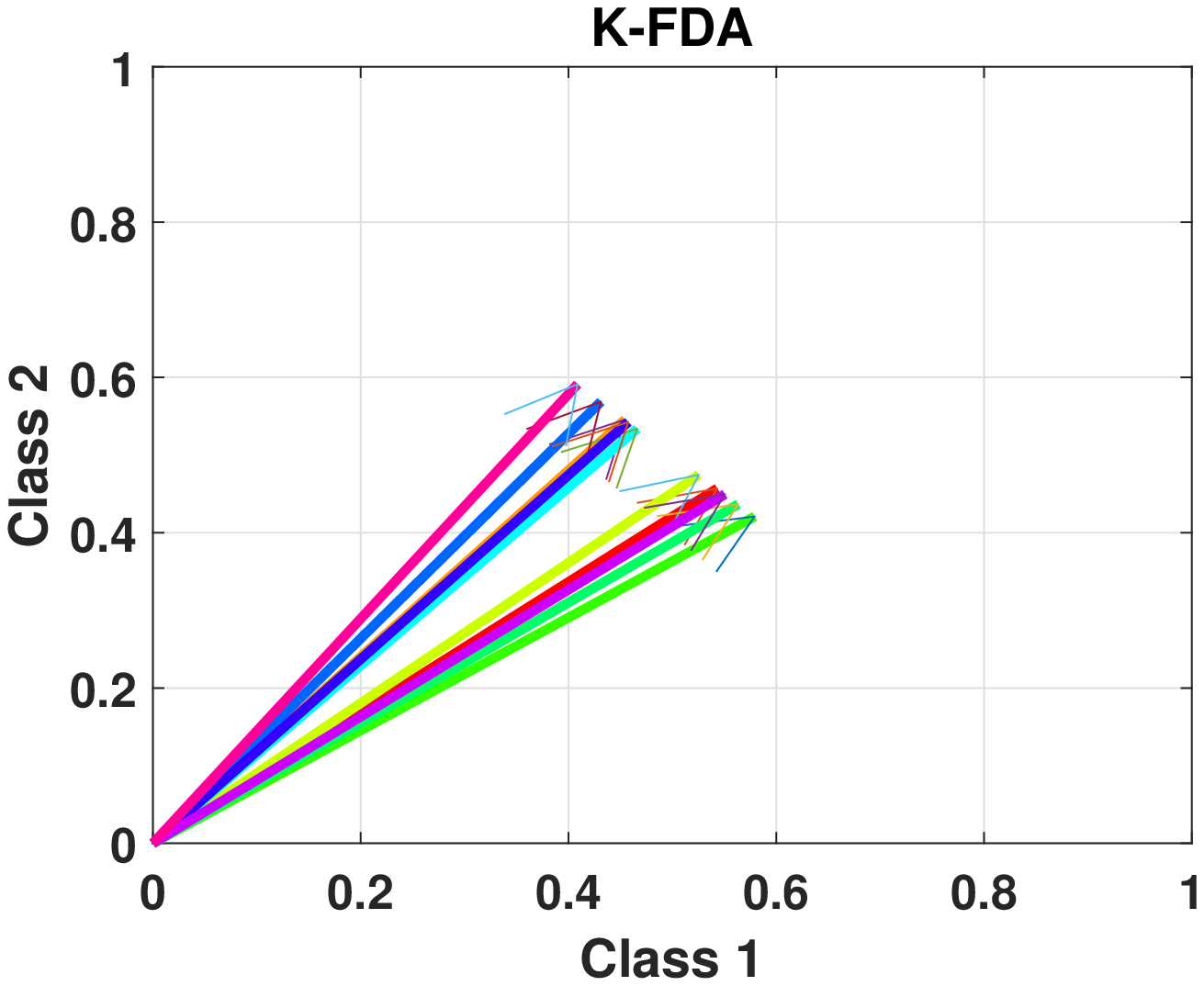}
	\end{subfigure}
	\begin{subfigure}{0.24\textwidth}
		\includegraphics[width=1\textwidth]{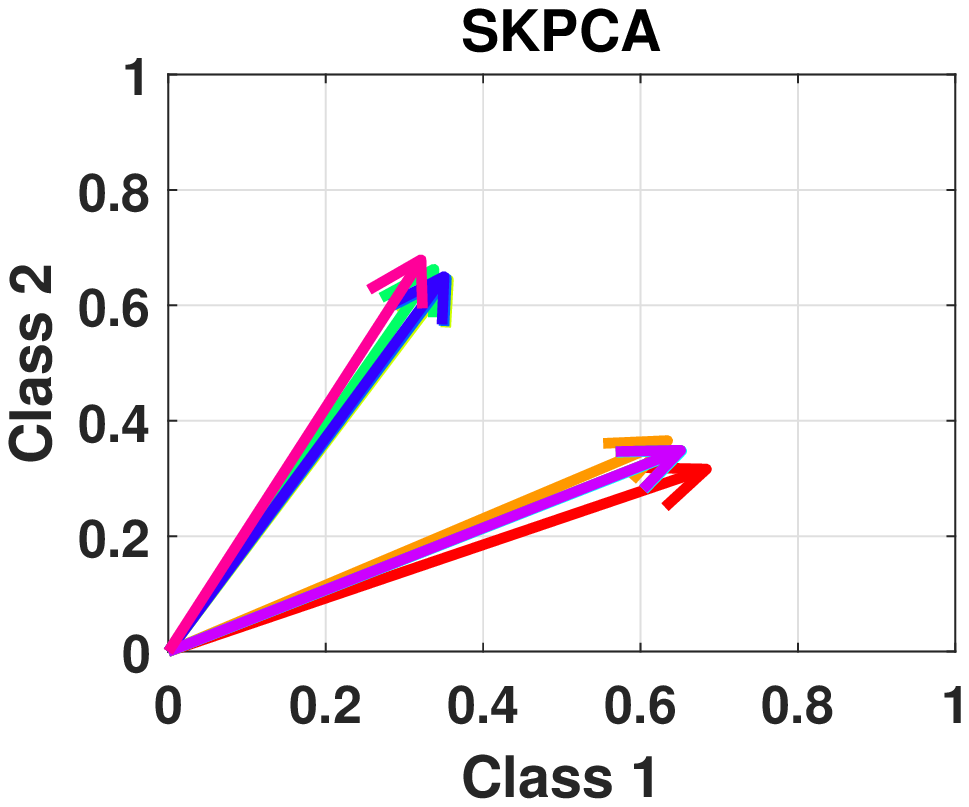}
	\end{subfigure}
	\begin{subfigure}{0.24\textwidth}	
		\includegraphics[width=1\textwidth]{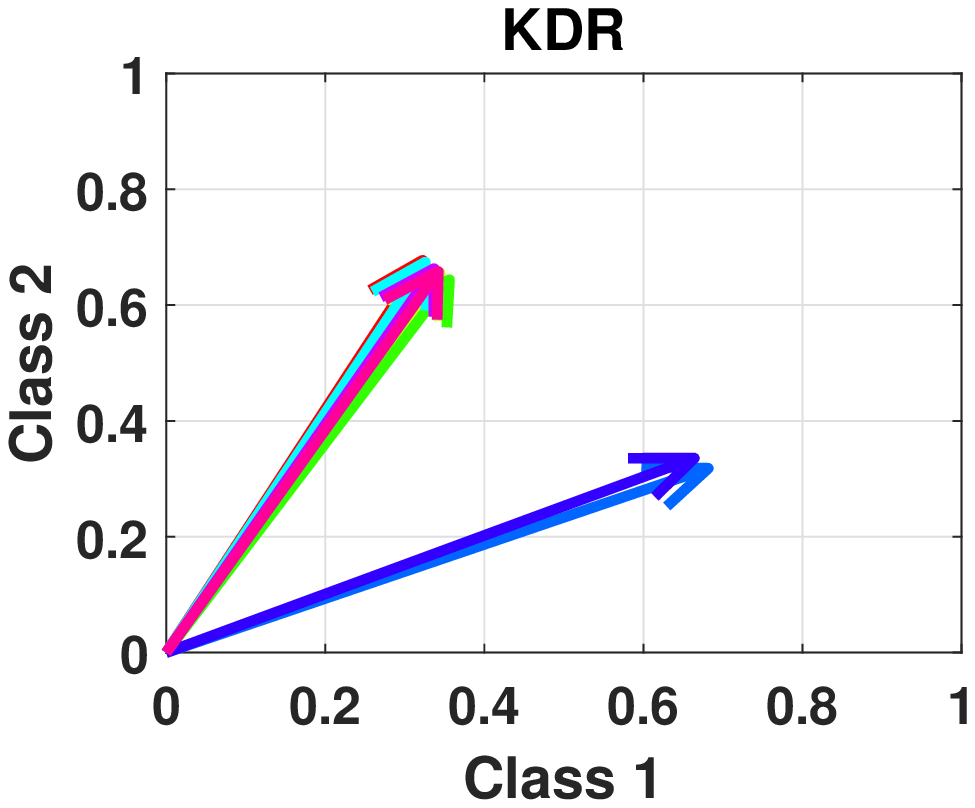}
	\end{subfigure}					
	\caption{Projecting the embedding dimensions on the label-space for the Sonar dataset.}
	\label{fig:HA}
\end{figure}

Additionally, to visualize the interpretation of the embeddings, we project the embedding dimensions on the label-space by computing $\MB{D}=\nH\nA \in \mathbb{R}^{\ncl\times k}$. Each column of $\MB{D}$ is a $\ncl$-dimensional vector that its $q$-th entry explains how strong is the relation of this dimension to the class $q$.
Figure~\ref{fig:HA} visualizes the columns of $\MB{D}$ for I-KDR, K-FDA, SKPCA, and KDR according to their implementations on the Sonar dataset. Each embedding was done for 10 target dimensions. Based on the results, I-KDR's embedding dimensions are almost separated into two distinct groups each of which mostly related to one class in the data. Although for SKPCA and KDR the vectors almost belong to two separate groups, they cannot be assigned to any of the classes confidently. For K-FDA, almost none of the above can be observed.

\subsection{Feature Selection}
In order to evaluate the feature selection performance of our I-KDR algorithm, we compute Eq.~(\ref{eq:gausker}) for each dimension of the data individually which results in a set of kernels $\{\K_i\}_{i=1}^\nf$ for each dataset.
We feed these kernels to the optimization framework of Eq.~(\ref{eq:relaxed}) to optimize their corresponding weights in $\nAlf$.
Besides the classification accuracy, we also measure $\|\nAlf\|_0$ to evaluate the feature selection performance of the algorithms.
Accordingly, we choose the following relevant set of baselines: 
MKL-TR~\cite{jiang2014trace},
MKL-DR~\cite{lin2011multiple},
KNMF-MKL~\cite{gu2015multiple},
and DMKL~\cite{wang2016discriminative}.

Based on Table~\ref{tab:FS},	
by optimizing the value of $\nAlf$ in Eq.~(\ref{eq:DR_frame}), I-KDR achieves better discriminations in the embedded space.
Consequently, as a general trend among the datasets, I-KDR's accuracies are improved after we optimized it in the multiple kernel framework (Compared to Table~\ref{tab:DR}).
Regarding the number of selected features, I-KDR, MKL-TR, and DMKL obtained similar results. 
Even more, for some of the datasets, the baselines obtained sparser feature selections than I-KDR.
Nevertheless, I-KDR demonstrates that its number of selected features are more efficient than others due to its supremacy in classification accuracies.
Therefore, we can claim that I-KDA performed more efficient than others in discriminative feature selection scenarios.
For CNS and Sonar dataset, I-KDR obtains competitive accuracy and feature selection performance compared to MKL-TR and DMKL, 
while for the Colon dataset, it outperforms the next best method (MKL-TR) with $7.73\%$ accuracy margin. 
As an explanation regarding the relatively high values of $\|\nAlf\|_0$ for KNMF-MKL, this algorithm uses a DR model, but it does not have a discriminative objective in its optimization.

\begin{table}[!h]
	\centering
	\footnotesize
	\caption{Comparison of classification accuracies ($\%$) and $\|\nAlf\|_0$ (in parenthesis).}	
	\begin{tabular}{l|c|c|c|c|c} %
		\hline	
		Dataset &I-KDR & DMKL & MKL-TR&MKL-DR & KNMF-MKL\\
		\hline
		\hline
		Yale&\textbf{83.22 (20)}&	78.25 (39)&	79.88 (34)&70.34 (93)&	68.43 (543)\\
		Sonar&87.91 (37)	&87.53\textbf{ (34)}&	87.94 (41)&70.34 (93)&	68.43 (543)\\
		Colon&\textbf{89.29} (25)	&80.32\textbf{ (21)}&	81.56 (34)&80.67 (67)	&78.43 (1321)\\
		20NG&\textbf{88.41} (73)	&85.01 (57)	&84.42\textbf{ (55)}&86.24 (384)&	83.11 (14483)\\				
		Gli85&\textbf{79.65 (33)}	&73.13 (54)&	74.46 (50)&72.83 (79)	&71.78 (10764)\\
		CNS &76.53 (47)	&76.37\textbf{ (32)}	&75.84 (25)&74.23 (109)	&72.43 (4872)\\
		Dbwork&\textbf{91.98 (29)}	&87.23 (41)	&86.53 (46)&85.14 (85)&	85.34 (1049)\\
		XM2VTS50&\textbf{97.74 (17)}&	92.76 (31)	&93.84 (29)&92.88 (55)&	90.89 (389)\\
		\hline
	\end{tabular} 			
	\\
	\scriptsize{The best result (\textbf{bold}) is according to a two-valued t-test at a $5\%$ significance level.}
	\label{tab:FS} 
\end{table}
\section{Conclusion}
In this paper, we proposed a novel algorithm to perform discriminative dimensionality reduction on the manifold.
Our I-KDR method constructs its embedding dimensions by 
selecting data points from local neighborhoods in the RKHS.
This strategy results in embeddings with better class-based interpretations for their bases.
Besides, by focusing on within-class local similarities and between-class dissimilarities, 
our method improves the separation of the classes in the projected space.
The I-KDR algorithm has a bi-convex optimization problem, and we use the alternating optimization framework to solve it efficiently.  
Furthermore, our approach can fuse the feature selection and dimensionality reduction for RKHS.
Our empirical results show that I-KDR outperforms other relevant baselines in both DR and feature selection scenarios.

\section*{Acknowledgement}
This research was supported by the Center of Cognitive 
Interaction Technology 'CITEC' (EXC 277) at Bielefeld University, which
is funded by the German Research Foundation (DFG).
\bibliographystyle{unsrt}
\bibliography{c:/Thesis/Publications/Ref4Papers_CS}



\end{document}